%% file: manuscript.tex
\documentclass[sigconf,nonacm]{acmart}
\AtBeginDocument{%
  }

\setcopyright{acmlicensed}
\copyrightyear{2026}
\acmYear{2026}
\acmDOI{XXXXXXX.XXXXXXX}
\acmConference[ICAIL'26]{21th International Conference on Artificial Intelligence and Law}{June 8--16, 2026}{Singapore}
\acmISBN{978-1-4503-XXXX-X/2018/06}

\usepackage{todonotes}
\usepackage{listings}
\usepackage{xcolor}

\begin{document}
%
%

\title{GDPR Auto-Formalization with AI Agents and Human Verification}


\author{Ha Thanh Nguyen}
\orcid{0000-0003-2794-7010}
\affiliation{%
  \institution{Center for Juris-Informatics, ROIS-DS}
  \city{Tokyo}
  \country{Japan}
}

\author{Wachara Fungwacharakorn}
\orcid{0000-0001-9294-3118}
\affiliation{%
  \institution{Center for Juris-Informatics, ROIS-DS}
  \city{Tokyo}
  \country{Japan}
}

\author{Sabine Wehnert}
\orcid{0000-0002-5290-0321}
\affiliation{%
  \institution{Ruhr-University Bochum, RC-Trust}
  \city{Bochum}
  \country{Germany}
}

\author{May Myo Zin}
\orcid{0000-0003-1315-7704}
\affiliation{%
  \institution{Center for Juris-Informatics, ROIS-DS}
  \city{Tokyo}
  \country{Japan}
}

\author{Yuntao Kong}
\orcid{0009-0001-2089-2363}
\affiliation{%
  \institution{Center for Juris-Informatics, ROIS-DS}
  \city{Tokyo}
  \country{Japan}
}

\author{Jieying Xue}
\orcid{0009-0000-8070-6609}
\affiliation{%
  \institution{Center for Juris-Informatics, ROIS-DS}
  \city{Tokyo}
  \country{Japan}
}

\author{Michał Araszkiewicz}
\orcid{0000-0003-2524-3976}
\affiliation{%
  \institution{Uniwersytet Jagielloński w Krakowie}
  \city{Kraków}
  \country{Poland}
}


\author{Randy Goebel}
\orcid{0000-0002-0739-2946}
\affiliation{%
  \institution{Alberta Machine Intelligence Institute, University of Alberta}
  \city{Edmonton}
  \state{Alberta}
  \country{Canada}
}

\author{Ken Satoh}
\orcid{0000-0002-9309-4602}
\affiliation{%
  \institution{Center for Juris-Informatics, ROIS-DS}
  \city{Tokyo}
  \country{Japan}
}

\renewcommand{\shortauthors}{Nguyen Ha Thanh et al.}

\begin{abstract}
We study the overall process of automatic formalization of GDPR provisions using large language models, within a human-in-the-loop verification framework. 
Rather than aiming for full autonomy, we adopt a role-specialized workflow in which LLM-based AI components, operating in a multi-agent setting with iterative feedback, generate legal scenarios, formal rules, and atomic facts. This is coupled with independent verification modules which include human reviewers' assessment of representational, logical, and legal correctness.
Using this approach, we construct a high-quality dataset to be used for GDPR auto-formalization, and analyze both successful and problematic cases. Our results show that structured verification and targeted human oversight are essential for reliable legal formalization, especially in the presence of legal nuance and context-sensitive reasoning.

\end{abstract}





\maketitle

\section{Introduction}

Recent advances in large language models (LLMs) have made it increasingly feasible to generate structured representations from natural language at scale. In legal informatics, this has renewed interest in legal formalization: translating legal norms into machine-executable representations for automated reasoning and compliance assessment. The General Data Protection Regulation (GDPR) has emerged as a key test case due to its complexity and practical relevance~\cite{zin2026can}.

However, the main challenge is not generating syntactically valid rules, but ensuring \emph{legal faithfulness}. Legal norms involve open-textured concepts, conditional obligations, role-dependent permissions, and a distinction between substantive rights and procedural safeguards. As a result, outputs that are logically consistent may still be legally invalid or misleading.

We argue that full autonomy is not appropriate for this task. Instead, legal formalization requires a verification-centered approach, where generation a formal representation is treated as provisional and correctness is established through structured scrutiny.

To this end, we propose a human-in-the-loop, multi-agent framework. A drafting agent generates scenarios, rules, and facts, while multiple verifier agents independently evaluate representational, logical, and legal correctness. This role separation is designed to reduce both bias and surface disagreements.

Using this framework, we construct a curated dataset for GDPR auto-formalization through iterative refinement, automated verification, and selective human validation.\footnote{The dataset is available at \url{https://huggingface.co/datasets/nguyenthanhasia/gdpr-cases}.} Our analysis shows that LLMs perform well on trigger-based legal conditions but fail on absolute rights, narrowly scoped exceptions, omission-based violations, and procedural–substantive distinctions. These failures arise primarily from abstraction choices rather than logical errors.

\textbf{Contributions.} (1) A verification-centered framework for legal formalization using role-specialized AI agents; (2) empirical insights into failure modes of AI-generated legal formalizations; and (3) a curated dataset for GDPR auto-formalization to support future research.

\section{Related Work}
This section reviews prior work in three closely related areas: legal formalization, the automation of formalization using AI techniques, and prior efforts to formalize the GDPR. While each line of research contributes important foundations, none directly addresses the problem of reliably validating automatically generated legal formalizations for complex regulations such as the GDPR.

\subsection{Legal Formalization}

Legal formalization serves as the necessary initial step for automated legal reasoning, by aiming to translate complex legal texts into machine-executable logical representations while preserving legal meaning.
 Theoretical research in this domain spans historical legal philosophy and modern computational logic. Historically, the debate over Legal Formalism has shaped the understanding of law as a system of abstract concepts subject to processing by deduction, which is in contrast with later Realist critiques that highlighted the social and policy choices inherent in legal interpretation. This tension underpins the modern challenge of formalizing laws \emph{open texture} \cite{hart2012concept} and continues to shape contemporary approaches to computational legal formalization.

Early computational work on legal formalization focuses on building legal expert systems based on logic programming languages, such as \emph{PROLOG} \cite{sergot_british_1986,sherman1987prolog}. 
Foundational research often focuses on adopting various non-classical logics, such as \emph{Deontic Logic}~\cite{von1951deontic,jones1992deontic} (to express obligations and permissions) and \emph{Defeasible Logic}~\cite{nute2001defeasible,prakken2013logical} (to manage rule conflicts and exceptions common in legal texts).

Current practice often integrates these logics within structured frameworks like \emph{LegalRuleML} \cite{lam2019enabling} and leverages domain ontologies to supply a shared vocabulary of legal concepts that supports rule formalization. Another approach intends to make representations more accessible to legal scholars. For example, \emph{PROLEG} \cite{satoh2010proleg}, derived from PROLOG, aims to separate negation-as-failure from exceptions, in order to mirror the legal practice of switching the burden of proof. Other approaches involve designing new programming languages for representing legal rules, such as \emph{Catala} \cite{merigoux2021catala}, or using controlled natural languages, such as \emph{Attempto Controlled English} \cite{fuchs2008attempto,wyner2015language} and \emph{Logical English} \cite{kowalski2020logical,kowalski2022logical}. These languages restrict their syntax so that sentences translate unambiguously into executable logical forms. However, current practices on legal formalization still rely on manual or semi-manual processes, and while being rigorous, they are often cited as time-consuming and expertise-reliant. Addressing this challenge requires both advanced information extraction techniques \cite{premasiri2025survey} and careful consideration of regulatory trade-offs;  this motivates our automated yet human-verified approach.

While these approaches provide expressive and legally informed formalisms, they largely presuppose that the formalization itself is correct. They offer limited guidance on how to validate formal representations that are generated automatically, particularly when errors arise from abstraction choices, scope misalignment, or subtle misinterpretations of legal conditions and exceptions.

\subsection{Automating Formalization}


AI researchers have long been seeking to automate the complex translation from natural language law to formal logic to overcome the labor-intensive nature of manual formalization. In semantic parsing, early systems like CHILL used inductive logic programming to learn mappings from sentences to Prolog queries \cite{DBLP:conf/aaai/ZelleM96}, which demonstrated feasibility but revealed limited scalability beyond narrow domains. Grammar-based approaches, such as those by \cite{DBLP:conf/uai/ZettlemoyerC05} improved compositional generalization but required logical form annotations.
Neural semantic parsers introduced encoder-decoder architectures that translate text to structured logical forms such as lambda calculus or Prolog-like representations  \cite{DBLP:conf/acl/DongL16}. While effective on benchmark datasets \cite{DBLP:conf/acl/LiangBLFL17}, these models were not designed for domains which require precise symbolic reasoning, such as general legal reasoning \cite{DBLP:conf/ijcai/AlvianoGSR25}.

Recent advances have delivered further leverage  with large language models (LLMs), which can be used to generate logic programs (e.g., Datalog \cite{alviano2025datalog}, ASP \cite{DBLP:conf/ijcai/AlvianoGSR25}, FOL \cite{DBLP:conf/acl/YangXPSF24,DBLP:conf/naacl/Liu25}) via prompting or fine-tuning \cite{DBLP:conf/fmcad/MendozaHT24,DBLP:conf/naacl/Liu25}.  Other results employ specialized techniques like Chain-of-Instruction (CoI) prompting to guide the generative process toward structured logical outputs, such as those in Defeasible Deontic Logic (DDL) \cite{horner2025legal}. A noted critical challenge is the scarcity of annotated legal datasets (informal-formal pairs), which has guided researchers to explore methods like back-translation and model specialization to generate high-quality training data, or to use LLMs in zero-shot or few-shot settings \cite{breton2025leveraging}. Some studies also show how LLMs can produce executable logic from natural language queries with high fluency, though often at the cost of formal correctness \cite{DBLP:conf/fmcad/MendozaHT24, alviano2025datalog}.

To address these challenges, hybrid neuro-symbolic approaches integrate LLMs with logic engines. For instance, some systems in legal reasoning use LLMs to extract candidate rules in Prolog, validated via symbolic execution for soundness and consistency \cite{DBLP:journals/corr/abs-2502-17638}.
These approaches increasingly treat logic generation as an interpretable intermediate layer, which should enable verifiable reasoning. By combining LLMs' linguistic flexibility with logic solvers' precision, current methods aim to bridge natural language and formal rule-based representations.

Despite these advances, most automated formalization approaches emphasize syntactic validity, logical consistency, or execution correctness. They rarely address whether a generated rule faithfully captures the normative scope of a legal provision, whether exceptions are applied too broadly, or whether procedural requirements are mistakenly encoded as conditions that negate substantive rights. Within this perspective, verification is thus treated primarily as a technical problem, rather than a legal one.

\subsection{GDPR Formalization}

The General Data Protection Regulation (GDPR) is often used in legal formalization studies, driven by the need for automated compliance checking, risk assessment, and enforcement of data subject rights. 
The formalizataion of GDPR provides the challenge of a highly structured, multi-layered approach to translating complex, principle-based legislation in GDPR into actionable logic. Research in this area is anchored by the creation of specialized, large-scale knowledge bases, such as DAPRECO \cite{robaldo2020formalizing}. This repository contains hundreds of GDPR rules formalized using sophisticated logics, often Reified Input/Output (RIO) Logic—an extension of I/O logic designed to handle the complex deontic and structural aspects of legal norms—and encoded using the standardized XML formalism, LegalRuleML. This work showcases the effective, manual application of logic to model permissions, obligations, and prohibitions across the breadth of the regulation.

A key enabler of this work is an ontology-level conceptualization of GDPR entities and relations. In particular, PrOnto (Privacy Ontology) \cite{palmirani2018pronto} models core GDPR concepts (e.g., data categories, processing purposes, legal bases, and roles such as Controller and Processor) and provides the semantic grounding needed to interpret and maintain large rule sets consistently.

However, existing GDPR formalization efforts are largely manual and expert-driven. While they provide authoritative examples of what correct formalization looks like, they do not address how such formalizations can be obtained reliably through automated or semi-automated processes. In particular, they do not consider how LLM-generated rules should be validated, how human oversight should be integrated, or how disagreements and ambiguities should be systematically analyzed during dataset construction.

\section{Task Definition and Formalism}

Here we define our auto-formalization task and introduce the formal representation used for legal rules and reasoning. Our purpose is to make explicit \emph{what} is being generated, \emph{what} is being evaluated, and under which semantic assumptions correctness is assessed, independently of the generation pipeline described below.

\subsection{Task Definition}

We pursue the task of GDPR auto-formalization coupled with human expert verification. Given the natural-language text of a GDPR article, the objective is to construct structured, machine-executable representations that support legal reasoning while remaining faithful to the normative content of the provision.

Each data sample consists of: (i) a \textbf{scenario} describing a concrete factual situation with a binary legal question, (ii) a set of atomic \textbf{facts} extracted from the scenario, (iii) a \textbf{rule tree} representing the legal norm, and (iv) a boolean \textbf{label} obtained by evaluating the rule against the facts.

The task goes beyond generating executable logic: a formalization is considered correct only if it preserves the scope of the legal norm, respects condition–exception structures, and does not conflate procedural safeguards with substantive rights.

Legal rules are represented using the Pythen framework~\cite{pythen}, a Python-based formalism for defeasible legal reasoning. Rules are encoded as condition–exception structures (``rule trees'') and evaluated against extracted facts to produce a boolean outcome.
We treat Pythen as an executable representation layer that enables systematic evaluation and verification. 

\subsection{Semantics and Evaluation Assumptions}

Rule evaluation is performed by a deterministic \textbf{RuleTreeEvaluator}, which computes the truth value of the target predicate by recursively traversing the rule tree and matching conditions against the provided facts.

The evaluation procedure follows these assumptions:
\begin{itemize}
    \item A closed set of facts is assumed for each scenario; predicates not present in the fact set are treated as not established.
    \item Exceptions have priority over conditions: if any exception predicate is satisfied, the rule is defeated regardless of satisfied conditions.
    \item Negation-as-failure is used implicitly for predicates that are not provable, unless explicitly represented as classical negation in the rule tree.
\end{itemize}

The resulting boolean label reflects the outcome of applying the formalized rule to the extracted facts under these assumptions. Crucially, note that a positive outcome \emph{does not necessarily imply legal compliance in practice.} For rights-based provisions, this may indicate only the existence of a right, independent of whether the controller subsequently complies. For permissibility-based provisions, the boolean leable expresses permissibility under the modeled conditions and exceptions, rather than a holistic assessment of compliance across the entire regulation.

These semantic choices are deliberately simple and transparent. Rather than aiming for full legal completeness, they provide a stable and interpretable foundation for studying where automated formalization succeeds or fails, and for subjecting generated rules to both automated and human verification.

\section{Methodology}

Here we describe the verification-centered process by which GDPR auto-formalization samples are generated, evaluated, refined, and selectively retained. Rather than treating data generation as a purely generative task, the methodology emphasizes structured verification, role separation, and staged and bounded iterative refinement to ensure legal and logical reliability.

\begin{figure*}[ht]
    \centering
    \includegraphics[width=0.5\textwidth]{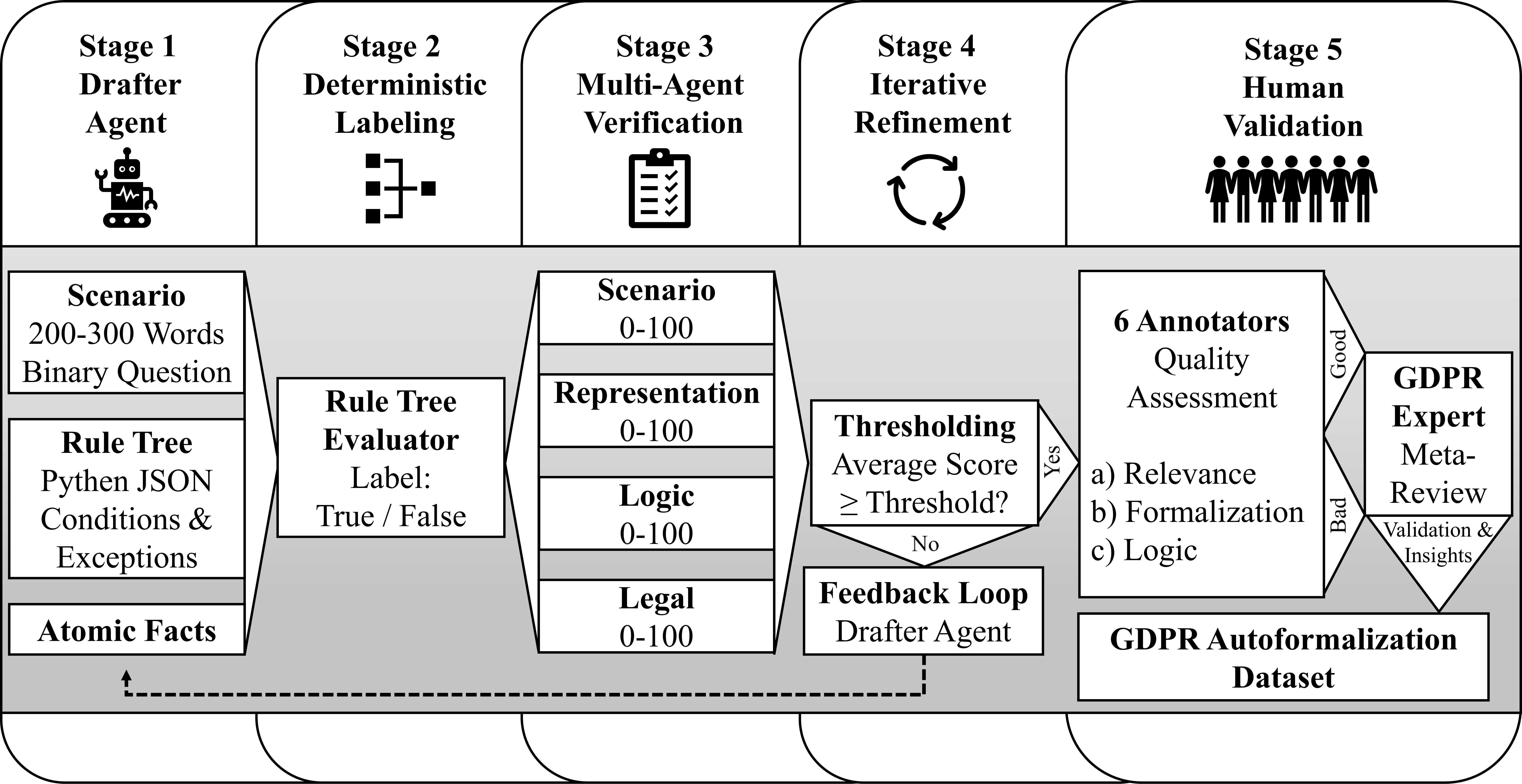}
    \caption{Overview of the verification-centered generation and validation pipeline.}
    \label{fig:rough_sketch}
\end{figure*}

The overall process employs five distinct AI agents: one \textbf{Drafter Agent} responsible for initial generation, and four \textbf{Verifier Agents} tasked with independent quality assessment. Automated verification is complemented by human validation and legal expert review, through which representative high-quality samples and systematic failure cases are identified and curated. The overview of the pipeline is demonstrated in Figure \ref{fig:rough_sketch}.

\subsection{Data Generation Pipeline}

Our data generation pipeline automates the transformation of GDPR articles into structured, machine-readable formalisms. This process consists of three primary stages: generation, automated verification, and human validation, operating in a feedback loop to progressively enhance data quality.

\subsubsection{Automated Content Generation}
The process is initiated by the Drafter Agent, which generates the core components of a sample given a target GDPR article:
\begin{enumerate}
    \item \textbf{Scenario Generation:} A natural-language narrative of approximately 200--300 words describing a concrete factual situation that falls within the scope of the target GDPR article and concludes with a binary (yes/no) legal question about the scenario.
    \item \textbf{Rule Formalization:} A formal representation of the legal norm encoded as a Pythen \texttt{rule\_tree}, capturing the condition-exception structure of the article.
    \item \textbf{Fact Extraction:} A set of atomic predicates extracted from the scenario, corresponding to leaf predicates in the rule tree and serving as inputs for evaluation.
\end{enumerate}

An illustrative example of a Pythen rule structure is shown in Listing~\ref{lst:pythen}. The example is schematic and intended to demonstrate the formal pattern rather than any specific GDPR provision.

\footnotesize
\begin{lstlisting}[language=Python, caption={Illustrative example of a Pythen rule structure.}, label={lst:pythen}]
{
    "p": "gdpr_art20_data_portability",
    "op": "ALL",
    "conditions": [
        "data_subject_requests_portability",
        "processing_is_automated",
        "processing_based_on_contract_or_consent"
    ],
    "exceptions": ["adversely_affects_others_rights"]
}
\end{lstlisting}
\normalsize
\subsubsection{Logical Rule Evaluation}
Following the generation of the scenario, rule tree, and facts, a deterministic \textbf{RuleTreeEvaluator} component then computes a derived label. This component functions as the execution engine for the Pythen framework. It systematically traverses the \texttt{rule\_tree}, evaluating the logical nodes by applying the set of extracted \texttt{facts}. The final output is a boolean label (\texttt{True} or \texttt{False}), representing the logical outcome of applying the formalized rule to the given facts.

This rule tree is generated specifically to operationalize and answer the yes/no question posed in the scenario; however, its structure also reflects that auto-formalization does not necessarily yield a one-to-one correspondence between the natural-language legal inputs—namely the scenario question and the governing legal norm—and their resulting logical representation.

\subsubsection{Multi-Agent Verification Framework}
To ensure the quality of the generated data, four specialized Verifier Agents perform automated evaluation, each focusing on a distinct analytical dimension. These agents provide quantitative scores and qualitative feedback:

\begin{itemize}
    \item \textbf{Scenario Verifier:} Evaluates the relevance, specificity, and realism of the generated scenario in the context of the target GDPR article. This agent assesses whether the scenario presents sufficient factual detail to enable a determinate legal conclusion.
    
    \item \textbf{Representation Verifier:} Validates the syntactic correctness and structural integrity of the \texttt{rule\_tree} according to the Pythen JSON schema. This agent verifies that all predicates are properly formed, that conditions and exceptions are correctly specified, and that the logical operators (ANY/ALL) are appropriately applied.
    
    \item \textbf{Logical Verifier:} Confirms that the label derived by the RuleTreeEvaluator is a logically sound consequence of the rules and facts. This agent independently re-evaluates the rule tree against the extracted facts to verify consistency.
    
    \item \textbf{Legal Verifier:} Conducts a substantive legal analysis of the sample. This involves assessing the degree to which the \texttt{rule\_tree} accurately captures the legal nuances of the GDPR article, the correctness and completeness of the extracted \texttt{facts}, and the overall validity of the legal reasoning implied by the sample.
\end{itemize}

Each Verifier Agent produces a numerical score (ranging from 0 to 100) as well as detailed feedback. The average score across all four agents determines the initial quality assessment of the sample.

\subsubsection{Iterative Refinement through Feedback}
The feedback from the four Verifier Agents is systematically passed back to the Drafter Agent. This information serves as a potential corrective input for the subsequent generation cycle. The Drafter Agent uses this feedback to adjust its generation strategy, thus aiming to improve the quality of the scenario, the precision of the rule formalism, and the relevance of the extracted facts. This iterative process continues until a generated sample achieves an average score above a predefined threshold, at which point it enters the human validation stage.

The threshold of 70 was initially selected as a pragmatic balance point where the averaged feedback from multiple independent Verifier Agents indicates sufficient structural correctness and relevance, while still allowing minor residual errors to be efficiently resolved during the subsequent human validation stage. This choice optimizes learning throughput without compromising overall quality control.

\section{Conlusions}
This paper presented a verification-centered, human-in-the-loop framework for GDPR auto-formalization using role-specialized LLM agents. The framework generates scenarios, Pythen rule trees, and atomic facts, and subjects them to independent automated verification and targeted human review, resulting in a curated dataset of GDPR auto-formalization samples.
Human evaluation shows that automated formalization is most reliable when legal norms can be expressed through explicit trigger conditions and when scenarios encode legally decisive facts with sufficient precision. More complex normative structures—such as absolute rights, narrowly scoped exceptions, omission-based duties, and role-dependent constraints—continue to require human verification.
Future work will prioritize article-level applicability checks, improved modeling of omissions and role separation, and broader coverage across GDPR provisions to further clarify the boundaries of reliable legal auto-formalization.

\section*{Acknowledgements}
This work was supported by the "R\&D Hub Aimed at Ensuring Transparency and Reliability of Generative AI Models" project of the MEXT, by JSPS KAKENHI Grant Numbers, 25H00522 and 25H01112, and by JST as part of Adopting Sustainable Partnerships for Innovative Research Ecosystem (ASPIRE), Grant Number JPMJAP25B2.

\bibliographystyle{ACM-Reference-Format}
\bibliography{bibliography}


\include{appendix}

\end{document}

%% file: appendix.tex
\appendix

\section{Human Verification and Detailed Findings}
\label{sec:appendix_human_verification}

The human verification stage was conducted by six annotators with prior academic and applied experience in artificial intelligence and legal reasoning. Each annotator evaluated a distinct subset of instances, with no overlap across annotators. As a result, inter-annotator agreement metrics were not computed, as the study design prioritized expert coverage and throughput over redundancy.

After the Drafter and Verifier Agents completed their iterative generation and verification cycle, a total of 400 samples exceeded the predefined quality threshold based on aggregated verifier scores. From this pool, human annotators further selected 120 samples that exhibited high confidence and minimal ambiguity in their factual and legal structure.

Each selected case includes the target GDPR article, a scenario description, a formalized rule tree, formalized facts, a labeled outcome, and feedback from verifier agents. The cases were evaluated according to several criteria: (a) whether the scenario is relevant to the target GDPR article; (b) whether, given the scenario and the article, the rules and facts are reasonably formalized; and (c) whether the resulting reasoning is logically sound.

The 120 samples were then categorized by the annotators into good cases and bad cases, reflecting clear alignment or misalignment with legal expectations. These categorized cases were subsequently reviewed by a legal expert acting as a meta-reviewer, who performed a second-round substantive legal verification.

\subsection{Overall Findings from Human Evaluation}

Human evaluation reveals a clear distinction between cases where automated formalization aligns with expert legal judgment and cases where it fails in structurally meaningful ways.

Across the reviewed cases, successful formalizations share a common pattern: the underlying legal norm can be expressed through explicit, trigger-based conditions, and the scenario encodes legally decisive facts with sufficient precision. In such cases, the generated rule trees correctly separate the existence of rights or obligations from subsequent compliance behavior, and the resulting reasoning aligns with statutory requirements. This pattern is observed in particular for provisions such as Article~20, where rights are defined by a finite set of cumulative conditions, and for Article~32, with scenarios that explicitly encode ongoing security obligations rather than static safeguards.

By contrast, problematic cases consistently involve legal structures that resist straightforward condition-exception modeling. These include absolute or non-derogable rights, cumulative disclosure duties, omission-based violations, and risk-calibrated obligations. In these settings, the automated formalizations tend to become over-permissive, either by implicitly balancing interests where the law does not permit balancing, by failing to represent legally operative absences as facts, or by collapsing distinct legal roles and purposes into a single abstraction.

\emph{A recurring source of error lies in abstraction choices rather than syntactic or logical flaws}. Several cases exhibit internally coherent and executable rule trees that nonetheless misrepresent the legal scope of exceptions, conflate procedural safeguards with substantive entitlement, or allow lawful processing by one actor to incorrectly justify downstream processing by another. These failures are not reliably detectable through automated verification alone and require human interpretation to identify.

In addition to substantive misalignment, human review also highlights inefficiencies in formalization, such as unnecessary rule chains and incorrect treatment of negation, which do not always affect the final outcome but can undermine interpretability and proper burden allocation. Taken together, these findings indicate that automated legal formalization performs reliably only within a constrained subset of legal structures, and that human verification remains essential for identifying norm-level misrepresentations that arise from abstraction, omission, or scope leakage.

\subsection{Qualitative Analysis Summary}

The detailed qualitative analysis of both successful ("Good Cases") and problematic ("Bad Cases") formalizations is provided in Appendix~\ref{sec:appendix_cases}. The analysis demonstrates that the automated formalization matches expert legal judgment when the underlying legal norm can be expressed through explicit, trigger-based conditions (e.g., a clean Article 20 portability case). However, failures often occur due to abstraction choices, such as controller-agnostic modeling, invalidation of hidden nuances in straightforward cases, misplacement of procedural identity verification as a condition, inefficiencies in formalized rule trees, and fact extraction anchored to the original setting rather than the proposed change. These findings underscore the necessity of human verification for identifying norm-level misrepresentations.

\section{Detailed Qualitative Analysis of Formalization Cases}\label{sec:appendix_cases}

\subsection{Illustrative Good Cases}
The following examples illustrate cases where the automated formalization matches expert legal judgment. These are not artificially simple scenarios, but realistic situations in which the system correctly models both the logical structure and the legal limits of GDPR provisions. In particular, they demonstrate appropriate abstraction choices, a clear distinction between the existence of rights and subsequent compliance behavior, and accurate representation of legal conditions and exceptions.
\subsubsection{A Clean Article 20 Portability Case}

\paragraph{Case description.}
Sofia is a paying subscriber of PulsePath, a fitness-tracking application operated under a monthly subscription contract. Upon account creation, she provided basic account data (name, email, age, height, weight) and connected her smartwatch. PulsePath processes these data by automated means in its cloud infrastructure, including manually entered workout and nutrition logs as well as observed data such as step counts, heart rate, sleep duration, GPS running routes, and workout timestamps. In addition, PulsePath generates training readiness scores and suggested workout plans using proprietary algorithms.

When deciding to switch to a competing service, EnduroCloud, Sofia submits a written request to PulsePath invoking her right to data portability under GDPR Article~20. She asks (i) to receive her personal data in a structured, commonly used, machine-readable format (e.g., CSV or JSON), and (ii) to have the same dataset transmitted directly to EnduroCloud via an available API connection, providing the endpoint and authorization token. PulsePath responds that it can provide only screenshots or PDF summaries and refuses direct transmission, citing a lack of support for integrations with competitors. It also states, without further explanation, that training readiness scores and workout plans will be excluded as "internal analytics."

\paragraph{Formal rule structure.}
The agents model the existence of the Article~20 right using a concise rule that mirrors the statutory conditions: a data subject request, personal data relating to the requester, a processing basis in consent or contract, and automated processing. Exceptions are limited to the explicit legal carve-out for adverse effects on the rights and freedoms of others.

\footnotesize
\begin{lstlisting}[language=Python, caption={Core rule capturing the existence of the Article 20 data portability right.}]
{
    "p": "gdpr_art20_data_portability",
    "op": "ALL",
    "conditions": [
        "data_subject_requests_portability",
        "data_is_personal_data_of_subject",
        "processing_based_on_consent_or_contract",
        "processing_is_automated"
    ],
    "exceptions": ["adversely_affects_others_rights"]
}
\end{lstlisting}
\normalsize

\paragraph{Outcome.}
Based on the asserted facts—processing based on contract, automated means, and a valid portability request concerning Sofia’s personal data—the predicate \textit{gdpr\_art20\_data\_portability} evaluates to true. This reflects the correct legal position: Sofia holds the Article~20 right with respect to the personal data she provided and the data observed from her use of PulsePath, irrespective of PulsePath’s refusal to comply or its choice of output format.

The strongest aspect of the agents' auto-formalization in this case is the correct separation between the \emph{existence of the right} and the controller's subsequent compliance behavior. By grounding the right exclusively in the statutory conditions of Article~20, the model avoids collapsing non-compliant responses—such as offering only PDFs, refusing direct transfer, or making vague claims about "internal analytics"—into factors that negate the right itself.

\subsection{Illustrative Bad Cases}
The negative cases presented below were chosen because they expose failure modes that are difficult to detect through automated verification alone. In many instances, the generated rule trees are syntactically valid and internally coherent, and in some cases even produce legally correct outcomes. Nevertheless, expert review reveals deeper modeling errors arising from abstraction choices, scope misalignment, or conflation of procedural and substantive legal conditions. These cases are therefore illustrative not of random mistakes, but of structurally meaningful risks inherent in automated legal formalization.
\subsubsection{Failure of Controller-Based Modeling}

\paragraph{Case description.}
A German manufacturing company, SteelForm AG, introduces a mandatory occupational health monitoring program for machine operators after several workplace accidents. SteelForm contracts an external occupational health provider to conduct periodic medical assessments, including fatigue screening, musculoskeletal evaluations, and mental health questionnaires. The provider processes detailed health data under medical confidentiality and issues fitness determinations (e.g., fit, fit with restrictions, unfit).

In parallel, SteelForm integrates the provider's digital platform with its internal HR system to streamline workforce planning. As part of this integration, SteelForm requests access not only to the fitness determinations but also to underlying medical indicators, such as anxiety scores, medication usage, and risk flags generated during assessments. SteelForm argues that both entities are jointly responsible for workplace safety and therefore operate under a shared compliance framework.

\paragraph{Auto-formalized rule outcome.}
In the agent-generated formalization, the processing of health data is evaluated at the level of the overall program rather than per controller. Because the occupational health provider lawfully processes the data under Article~9(2)(h), the exception is treated as applying globally to the entire data flow. As a result, the model concludes that subsequent storage and use of detailed health data by SteelForm's HR department is permitted.

\paragraph{Modeling failure.}
This outcome is legally incorrect. Article~9(2)(h) authorizes processing by or under the responsibility of health professionals for occupational or preventive medicine purposes. It does not extend to employer-side HR processing for staffing, performance management, or operational planning. By collapsing the occupational health provider and the employer into a single undifferentiated controller context, the model erases the legal boundary that limits how far the exception can travel.

\paragraph{Analysis.}
The failure arises from treating "the program" as the unit of analysis instead of treating each controller and purpose separately. In legal reasoning, the permissibility of processing must be assessed independently for each controller, even where data originate from a lawful upstream activity. The auto-formalization fails to enforce this separation and instead allows a lawful medical exception to leak into non-medical HR decision-making. This results in a false positive: processing is labeled lawful despite the absence of explicit consent or any other applicable Article~9 exception for the employer.

\paragraph{Interpretive significance.}
This case illustrates a structural risk of controller-agnostic modeling. When controller boundaries are not represented as first-class elements in the rule system, lawful processing by one actor can incorrectly legitimize downstream processing by another. The error does not stem from misunderstanding Article~9's text, but from an abstraction choice that suppresses the controller dimension entirely. Human review is required to reintroduce this distinction and prevent exception scope expansion across organizational roles.

\subsubsection{Invalidation of Hidden Nuances in Straightforward Cases}

\paragraph{Case description.}
MediShip is a licensed online pharmacy in Germany that processes customer data for multiple purposes: fulfilling prescription orders, complying with statutory pharmacy and tax obligations, preventing fraud and securing its IT systems, and (optionally) sending promotional emails. In the described case, Anna orders a prescription asthma inhaler and provides the necessary personal and prescription data. While the processing required for fulfilment, legal compliance, and fraud prevention is clearly necessary and expected, the checkout flow presents a separate, unticked box for marketing emails, which Anna leaves unchecked. A subset of the facts, drafted by the Drafter Agent and having passed the review stage of the Verifier Agents, is shown in Listing~\ref{lst:mediship-marketing-facts}.

\paragraph{Formalized fact pattern (marketing purpose).}
\footnotesize
\begin{lstlisting}[language=Python, caption={Formalized facts for the marketing-email purpose in the MediShip case.}, label={lst:mediship-marketing-facts}]
[
    "necessary_for_legitimate_interests", 
    "necessary_for_contract_performance", 
    "necessary_for_legal_obligation", 
    "rights_not_overriding_legitimate_interests"
]
\end{lstlisting}

\normalsize

\paragraph{Analysis.}
At a purely syntactic level, it may appear straightforward to auto-formalize the marketing-email processing under GDPR Article~6(1)(f) by checking whether the controller asserts a legitimate interest and provides an opt-out mechanism. However, the natural-language nuances of the case invalidate this formal pattern. The facts explicitly state that marketing is not necessary for contract performance or legal compliance, and the unticked consent box signals that the data subject did not expect or agree to promotional use of her contact details. As a result, the predicate \textit{necessary\_for\_legitimate\_interests} is weakly supported at best, and the balancing predicate \textit{rights\_not\_overriding\_legitimate\_interests} is contradicted by the data subject’s expressed choice.

\subsubsection{Misplacement of Identity Verification as a Condition}

\paragraph{Case description.}
In the StrideSync scenario, Nina submits an Article~20 data portability request covering both receipt of her data in a machine-readable format and direct transmission to a competing controller. Substantively, the request concerns personal data relating to Nina, is processed by automated means, and is based on contract and consent. From a legal perspective, these elements place the request squarely within the scope of Article~20.

\paragraph{Excerpt from the formal rule model.}
\footnotesize
\begin{lstlisting}[language=Python, caption={Excerpt of the rule tree where identity verification is modeled as a condition.}]
{
    "p": "personal_data_concerns_requester",
    "op": "ALL",
    "conditions": [
        "identity_verified",
        "data_is_personal_data_of_requester"
    ],
    "exceptions": ["request_is_for_third_party_data"]
}
\end{lstlisting}
\normalsize

\paragraph{Formal modeling issue.}
Within the agent-generated rule tree, the predicate \textit{identity\_verified} is embedded as a mandatory condition for establishing that the requested data concerns the requester. While this reflects a legitimate procedural safeguard in practice, its placement in the logical structure causes the entire predicate \textit{personal\_data\_concerns\_requester} to evaluate to false whenever identity verification is treated as incomplete or ambiguous by the agent.

\paragraph{Analysis.}
This modeling choice illustrates how a procedural requirement can unintentionally flip the outcome of a rights analysis. In legal reasoning, identity verification functions as a precondition for execution and safe disclosure, not as a substantive limiter on the existence of the Article~20 right itself. By contrast, in the formal representation, failure to satisfy \textit{identity\_verified} collapses the higher-level predicate \textit{art20\_data\_portability\_right}, even when all substantive criteria are met. The result is a false negative: the right is treated as non-existent rather than temporarily pending verification. This demonstrates that auto-formalization, while seemingly straightforward, is sensitive to how procedural safeguards are encoded, and that human review is required to distinguish execution-related checks from conditions that genuinely negate a legal right.

\subsubsection{Inefficiencies in Formalized Rule Trees}

\paragraph{Analysis.} In some cases, although the generated rule trees and facts lead to legally correct outcomes, the rule trees themselves are sometimes inefficiently formalized. For example, the rule formalization agent sometimes fails to distinguish between classical negation and negation as failure, the latter of which is typically treated as an exception. As a result, this confusion can produce inefficient rule trees with unnecessary conditions, as illustrated in the following example:

\footnotesize
\begin{lstlisting}[language=Python, caption={Example of unnecessary conditions.}, label={lst:burden-problem}]
{
    "p": "no_art9_exception",
    "op": "ALL",
    "conditions": ["no_vital_interests_basis",
    "no_nonprofit_basis",
    "not_manifestly_public", 
    "no_legal_claims_basis", 
    "no_substantial_public_interest", 
    "no_medicine_basis", 
    "no_public_health_basis", 
    "no_research_archiving_basis"],
    "exceptions": []
},
\end{lstlisting}

\normalsize

Although the generated rule trees and facts can reach the legally correct outcome because the facts explicitly include all conditions appearing in the rules, such as \textit{no\_vital\_interests\_basis}. This formalization does not align with standard legal reasoning. In particular, \textit{no\_vital\_interests\_basis} should be assumed to hold by default unless its opposite, \textit{vital\_interests\_basis}, is established. Treating such negations as failure as explicit conditions with classical negations shifts the burden of proof incorrectly and results in unnecessary fact extraction, even though the same legal outcome could be achieved.

Meanwhile, the generated rule trees may contain unnecessary rule chains, as shown below:

\footnotesize
\begin{lstlisting}[language=Python, caption={Example of an unnecessary rule chain.}, label={lst:rule-chain}]
{
    "p": "no_a9_exception_applies",
    "op": "ALL",
    "conditions": ["no_a9_grounds_true"],
    "exceptions": []
},
{
    "p": "no_a9_grounds_true",
    "op": "ALL",
    "conditions": ["not_any_a9_ground_true"],
    "exceptions": []
},
{
    "p": "not_any_a9_ground_true",
    "op": "ALL",
    "conditions": [...],
    "exceptions": []
},
\end{lstlisting}

\normalsize

Listing~\ref{lst:rule-chain} shows that, in order to derive \textit{no\_a9\_exception\_applies}, the system must first prove the condition \textit{no\_a9\_grounds\_true}. To establish \textit{no\_a9\_grounds\_true}, it then needs to prove \textit{not\_any\_a9\_}
\textit{grounds\_true}. However, since there are no other rules that conclude \textit{no\_a9\_grounds\_true}, the intermediate rule introducing this predicate is redundant and can be eliminated without affecting the final outcome.

During the experiments, we also observed that agents sometimes hack the formalization to achieve the labeled outcomes. For example, the fact-extraction agent may extract no facts so that the derived reasoning does not entail an outcome labeled as false. Furthermore, we found that the logical-validation agent occasionally exceeds its time limit during validation, indicating that reasoning over complex logical structures remains challenging.

\subsubsection{Fact Extraction Anchored to the Original Setting Rather Than the Proposed Change}

\paragraph{Case description.}
BrightFit GmbH offers two optional consent-based purposes during onboarding, each controlled by an unticked toggle and accompanied by purpose-specific information. In the baseline design, users can proceed without enabling either toggle; consent is logged with contextual metadata (including the consent text version); and withdrawal is available via a simple settings control. The scenario then introduces a \emph{proposed change}: pre-ticking the advertising toggle, making withdrawal harder to locate, and weakening records by storing only a timestamp. The yes/no question asks whether \emph{these proposed changes} comply with GDPR Article~7.

\paragraph{Fact extraction inconsistency.}
Although the rule tree appropriately models Article~7 requirements and invalid consent patterns, the extracted facts primarily reflect the baseline compliant interface rather than the proposed non-compliant change (Listing~\ref{lst:change-problem}). In particular, predicates such as \textit{not\_default\_opt\_in} and \textit{consent\_record\_kept} remain present, while change-specific predicates that would activate the relevant Article~7 failures are absent (Listing~\ref{lst:change-correction}). This misaligns the operative state described by the question with the factual substrate used for evaluation.

\footnotesize
\begin{lstlisting}[language=Python, caption={Extracted facts reflect the original compliant interface, not the proposed changes.}, label={lst:change-problem}]
[
  "opt_in_action_recorded",
  "no_detriment_if_consent_refused",
  "separate_consent_for_separate_purposes",
  "consent_tied_to_specific_purpose",
  "not_default_opt_in",
  "data_subject_given_required_info",
  "consent_record_kept",
  "info_presented_clearly",
  "no_ambiguity_in_choice"
]
\end{lstlisting}

\begin{lstlisting}[language=Python, caption={Example facts needed to align the representation with the proposed change.}, label={lst:change-correction}]
[
  "consent_by_pre_ticked_boxes",
  "withdrawal_harder_than_giving",
  "insufficient_consent_record",
  "cannot_link_subject_purpose_time_to_text_version"
]
\end{lstlisting}
\normalsize

\paragraph{Analysis.}
The \texttt{False} label is correct for the proposed change: pre-ticked advertising consent violates the requirement of a clear affirmative action and triggers the pre-ticked-box invalidity; withdrawal friction conflicts with Article~7(3); and omission of the consent text version weakens demonstrability under Article~7(1). However, because the extracted facts do not encode these change-specific violations, evaluation can be driven toward the baseline configuration rather than the proposed change in question.